%% file: acl_latex.tex
\pdfoutput=1

\documentclass[11pt]{article}

\usepackage[preprint]{acl}

\usepackage{times}
\usepackage{latexsym}
\usepackage{multirow}

\usepackage[T1]{fontenc}

\usepackage[utf8]{inputenc}

\usepackage{microtype}

\usepackage{inconsolata}

\usepackage{graphicx}

\usepackage{pgfplots,pgfplotstable}
\tikzset{
  font={\fontsize{8pt}{10}\selectfont}}
\usepackage{graphicx}
\usepackage{subfigure}
\usepackage{arydshln}

\definecolor{darkyellow}{RGB}{251,188,4}
\definecolor{darkgreen}{RGB}{52,168,83}
\definecolor{lightblue}{RGB}{66,133,244}
\definecolor{acqua}{RGB}{70,189,198}

\title{SimulSeamless:\\ FBK at IWSLT 2024 Simultaneous Speech Translation}

\author{Sara Papi \and Marco Gaido \and Matteo Negri \and Luisa Bentivogli\\
Fondazione Bruno Kessler, Italy\\
\texttt{\{mgaido,spapi,negri,bentivo\}@fbk.eu} \\}

\begin{document}
\maketitle
\begin{abstract}
This paper describes the FBK's participation in the Simultaneous Translation Evaluation Campaign at IWSLT 2024. For this year's submission in the speech-to-text translation (ST) sub-track, we propose \textbf{SimulSeamless}, which is realized by combining AlignAtt and SeamlessM4T in its medium configuration. The SeamlessM4T model is used "off-the-shelf" and its simultaneous inference is enabled through the adoption of AlignAtt, a SimulST policy based on cross-attention that can be applied without any retraining or adaptation of the underlying model for the simultaneous task.
We participated in all the Shared Task languages (English$\rightarrow\{$German, Japanese, Chinese$\}$, and Czech$\rightarrow$English), achieving acceptable or even better results compared to last year's submissions. SimulSeamless, covering more than 143 source languages and 200 target languages, is released at \url{https://github.com/hlt-mt/FBK-fairseq/}. 
\end{abstract}

\begin{figure*}[!ht]
    \centering
    \includegraphics[width=\textwidth]{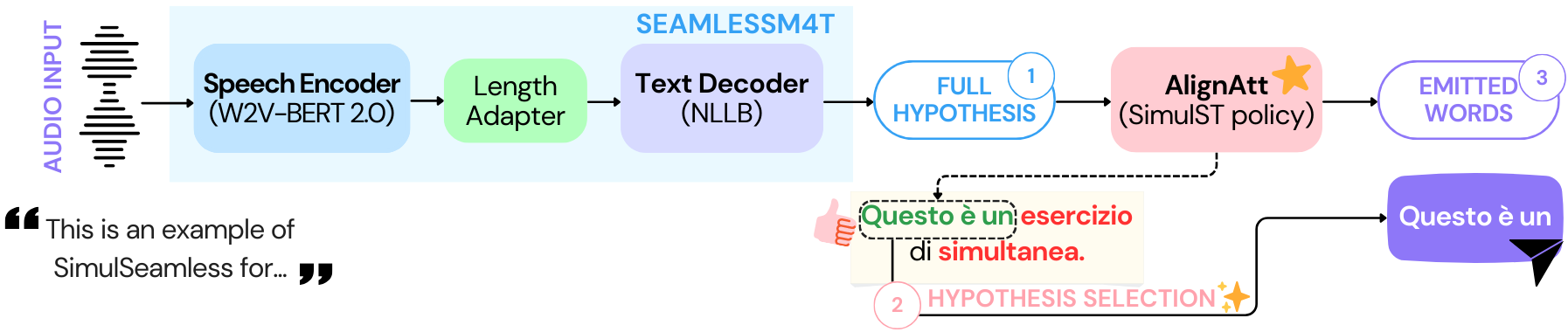}
    \caption{Representation of the SeamlessM4T model combined with AlignAtt: SimulSeamless.}
    \label{fig:simulseamless}
\end{figure*}

\section{Introduction}

Simultaneous speech-to-text translation (SimulST) is the task in which a model has to provide a textual translation into the target language while continuously receiving an incremental speech input in the source language. 

SimulST poses additional difficulties to standard offline ST, as it has to find the optimal balance between translation quality and output latency, which is the time delay between an utterance being spoken and the corresponding translation being emitted. This balance -- often referred to as "quality-latency tradeoff" -- depends on the application scenario \citep{fantinuoli-prandi-2021-towards}, which can span many domains such as online meetings, lectures, conference talks, and live shows. 

Due to the growing interest in SimulST technologies, this task has been included in the IWSLT Evaluation Campaigns\footnote{\url{https://iwslt.org/}} since 2020.
The increasing interest has led to numerous direct and cascade models participating in the challenge every year \citep{ansari-etal-2020-findings,anastasopoulos-etal-2021-findings,anastasopoulos-etal-2022-findings,agrawal-etal-2023-findings}, all vying for the title of the best approach to realize a SimulST system from scratch. More recently, the practice of using models without ad-hoc training for the simultaneous scenario has become widespread \citep{polak-etal-2022-cuni,gaido-etal-2022-efficient,papi-etal-2023-direct,polak-etal-2023-towards,yan-etal-2023-cmus,huang-etal-2023-xiaomi}, demonstrating that competitive or even superior results can be achieved compared to systems specifically tailored for SimulST \citep{papi2022does}. Among the strategies used to repurpose standard (offline) ST models for SimulST \citep{liu20s_interspeech,papi2022does,papi-etal-2023-attention}, AlignAtt \citep{papi-et-al-2023-alignatt} emerged as the best one, achieving new state-of-the-art results.
AlignAtt exploits speech-translations alignments based on cross-attention scores to guide the simultaneous inference, overcoming the limitations of the previous approach relying on attention \citep{papi-etal-2023-attention}.

Alongside the increased interest in the SimulST task, especially during the last year, we have witnessed an explosion in the use of large models \citep{latif2023sparks}, including speech foundation models \citep{pmlr-v202-radford23a,pratap2023mms,barrault2023seamlessm4t,zhang2023google}. These models are now commonly used alone or in combination with large language models \citep{gaido2024speech} for generic ST tasks. Among these, SeamlessM4T \citep{barrault2023seamlessm4t} has emerged as one of the most promising multimodal and multilingual models, covering more than 143 source languages and 200 target languages.  

For this year's submission to the IWSLT Evaluation Campaign on Simultaneous Translation, we, therefore, propose to combine the best of both worlds to obtain a multilingual model without any training or adaptation for the SimulST task. This results in \textbf{SimulSeamless}, consisting of the SeamlessM4T model used "off-the-shelf" repurposed for simultaneous inference using AlignAtt. 

From empirical results on the task, we show that SimulSeamless can achieve acceptable or even better results compared to last year's participants, despite not being retrained or fine-tuned either for the simultaneous task or on paired data in the evaluated languages. Moreover, SimulSeamless is a generic multilingual model that can be used for any allowed translation direction supported by the underlying SeamlessM4T model, covering more than 143 source languages and 200 target languages. 
The code is released under the Apache 2.0 Licence at \url{https://github.com/hlt-mt/FBK-fairseq/blob/master/fbk\_works/SIMULSEAMLESS.md}.

\section{SimulSeamless}
Similarly to previous years \citep{gaido-etal-2022-efficient,papi-etal-2023-direct}, we participated in the Simultaneous Translation evaluation campaign, focusing on the speech-to-text translation sub-track. For this year's submission, we opted for the use of the new SeamlessM4T model, which is allowed for the task,\footnote{\url{https://iwslt.org/2024/simultaneous}} as the underlying model of the SimulST policy AlignAtt. This policy can be applied to any standard (i.e., offline-trained) model without the need for retraining or adaptation.

In the following, both these elements and their combination are explained in detail.

\paragraph{SeamlessM4T.} SeamlessM4T \citep{barrault2023seamlessm4t} (or Massively Multilingual \& Multimodal Machine Translation) is a family of models based on pre-trained models including W2V BERT 2.0, and NLLB \citep{costa2022no}, whose encoder and decoder respectively are used for the speech-to-text modality. W2V-BERT is a Conformer-based model \citep{gulati20_interspeech} composed of 24 layers, with a total of $\sim$600M parameters, and trained on 1 million hours of open speech audio data to learn self-supervised speech representations. It processes the audio features obtained by applying 80-dimensional Mel filterbanks to the audio waveform. The W2V-BERT encoder is followed by a Length Adapter based on a modified version of the M-adaptor \citep{zhao22g_interspeech}, which is a Transformer-based model \citep{NIPS2017_3f5ee243} that is in charge of compressing the speech representation (by a factor of 8) through attention pooling. The compressed input representations are then fed to the NLLB decoder, in its 1.3B parameters configuration, to produce the translations. The final model was obtained after training on both manual and automatically aligned speech translation data with a total of 406,000 hours.

\paragraph{AlignAtt.} AlignAtt \citep{papi-et-al-2023-alignatt} is a SimulST policy that relies on cross-attention to make decisions about whether to emit translated words or wait for additional information in the simultaneous scenario. At each time step, the cross-attention scores are exploited to obtain audio-translation alignments by uniquely assigning the predicted words to the audio frames (encoder states) having the maximum attention score. Then, it is checked, for each word, if it has been aligned with one of the last $f$ frames, which is the parameter handling the latency of the model. If this is true, the emission is stopped, otherwise, the next word is evaluated. The idea behind AlignAtt is that, if a word is aligned with one of the last received audio frames, the encoded information could be unstable and/or not sufficient to reliably predict that word. Conversely, if a word mostly attends to a more stable and earliest-received encoded information, it can be safely predicted.
With this formulation, AlignAtt simplifies the previous EDAtt policy \citep{papi-etal-2023-attention} by eliminating the dependency on additional hyper-parameters while achieving competitive or even better results. 

\paragraph{SeamlessM4T + AlignAtt = SimulSeamless.} Since AlignAtt is applicable to any standard ST models without the need for re-training or adaptation, we chose to apply it directly to the SeamlessM4T model in its medium configuration, realizing \textbf{SimulSeamless}.
This solution
is completely different from SeamlessStreaming \citep{barrault2023seamless}, which is obtained through 
an expensive ad-hoc finetuning
of the Seamless model for the simultaneous task
based on
EMMA -- efficient monotonic multi-head attention \citep{ma2023efficient}.
Since
SeamlessM4T already covers all the languages evaluated in the Simultaneous track, the model is used completely "off-the-shelf". The SimulSeamless model is shown in Figure \ref{fig:simulseamless}.

\section{Experimental Settings}
We used the available checkpoint of the SeamlessM4T model provided on HuggingFace in its "medium" configuration,\footnote{\url{https://huggingface.co/facebook/seamless-m4t-medium}} with a total of 1.2B parameters.

The results are reported on the benchmarks used for the submission, which is MuST-C \citep{CATTONI2021101155} v2.0 tst-COMMON for en-\{de, ja, zh\}, and the dev set provided for the task for cs-en.
The scores are computed using the SimulEval toolkit \citep{ma-etal-2020-simuleval}.\footnote{We used the \texttt{f1f5b9a} commit that is the last version with the remove evaluation working, which is needed to run SimulEval using Docker containers.} 
Translation quality is evaluated using BLEU score with sacreBLEU \citep{post-2018-call}\footnote{Version 2.4.0.}. Latency is reported using Average Lagging (AL) \citep{ma-etal-2019-stacl} since it is the metric used for the final scoring. Length Adaptive Average Lagging (LAAL) \citep{papi-etal-2022-generation} and Average Token Delay (ATD) \citep{kano2022average} are also evaluated and included in the final results since they are official metrics reported for the task.\footnote{\url{https://iwslt.org/2024/simultaneous}} 
Both latency and BLEU scores are computed at the character level for Chinese and Japanese while the standard \texttt{13a} tokenizer is used for sacreBLEU, and word-level latency is computed for the other languages.
Additionally, computationally aware metrics are presented to account for the real elapsed time, which also considers the computational cost of running the underlying model. The inference was run using a single GPU NVIDIA V100 with 16GB of RAM. 

\begin{figure}[ht]
    \centering
    \includegraphics[width=0.475\textwidth]{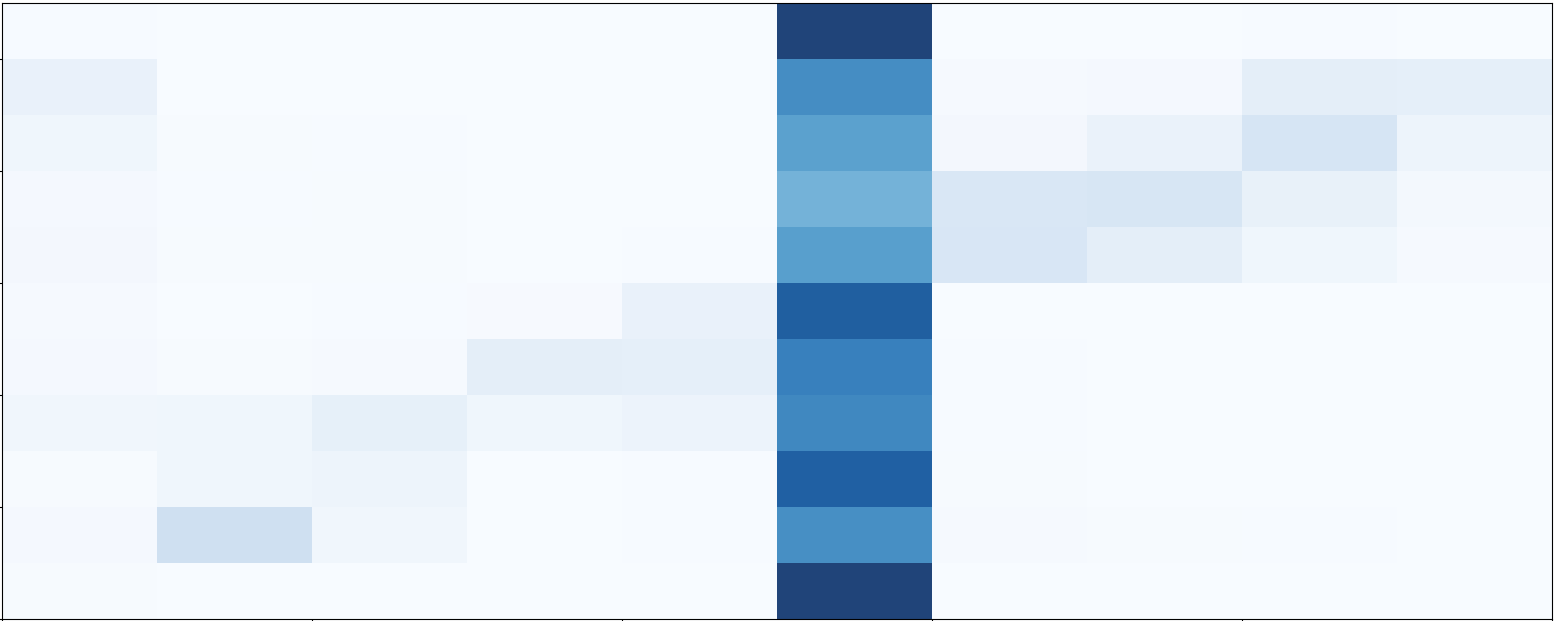}
    \caption{Example of skewed cross-attention scores representation towards some frames.}
    \label{fig:att-pattern}
\end{figure}

\input{de_tst_mustv2}

For the AlignAtt policy, we set the size of the speech chunk processed by the model at each time step to
1s for English to German and Czech to English, 800ms for English to Chinese, and 400ms for English to Japanese. To achieve latency close to an AL of 2s required for the submission, we set the hyper-parameter handling the latency $f$ to 1 for en-ja and en-zh, 6 for en-de, and 9 for cs-en. The cross-attention scores are normalized frame-wise before applying AlignAtt to avoid the cross-attention weights being skewed to some frame representation, as shown in Figure \ref{fig:att-pattern}.

\section{Results}

\subsection{Submission Selection}
For selecting the best setting, we analyzed the performance by varying the layer from which cross-attention scores are extracted since simply averaging them across layers led to worse results, as also already found in \citep{papi-etal-2023-attention}.
The layer-wise quality results are shown in Figure \ref{fig:bleu_res} while layer-wise latency results close to AL=2s are shown in Figure \ref{fig:al_res}. 

It can be seen from the Layer-AL(s) curves (Figure \ref{fig:al_res}) that Layer 5 represents a threshold layer starting from which the latency increases significantly without, however, similar significant quality improvements in terms of BLEU (Figure \ref{fig:bleu_res}). The only acceptable layers to achieve an AL$\leq$2s for en-ja are layers 1 and 2 while this set is extended to layer 4 for cs-en, and up to layer 5 for en-de and en-zh. Among the two admissible layers for en-ja, we chose for the final submission the one maximizing the quality, which is Layer 1. For en-zh, we followed a similar approach by choosing Layer 4, which achieves the highest BLEU score with an admissible latency. 
The choice of Layer 4 is also maintained for en-de and cs-en since we found that is the layer achieving the best quality-latency tradeoff between BLEU and AL.

\begin{table*}[!h]
  \centering
  \begin{tabular}{c||l|c|c|c|c}
    \hline
    \textbf{Lang. Pair} & \textbf{Model} & \textbf{BLEU $\uparrow$} & \textbf{AL $\downarrow$} & \textbf{LAAL $\downarrow$} & \textbf{ATD $\downarrow$} \\
    \hline
    & CMU$^\dag$ & 30.4 & 1.92 & 1.99 & - \\
    & CUNI-KIT$^\dag$ & 31.4 & 1.955 (3.072) & - & - \\
    & FBK$^\dag$ & 30.70 & 1.888 (2.939) & 2.069 (3.052) & 1.797 (2.364) \\
    & HW-TSC$^\ddag$ & 33.54 & 1.88 & - & - \\
    & NAIST & 29.98 & 1.964 & 2.173 & 1.894 \\
    \multirow{-6}{*}{en-de} & \textbf{SimulSeamless$^\dag$} & 27.37 & 1.815 (3.012) & 1.993 (3.137) & 1.778 (2.353) \\
    \hline
    & NAIST & 15.32 & 1.974 & 2.291 & 0.548 \\
    & CUNI-KIT$^\dag$ & 15.3 & 1.982 (3.489) & - & - \\
    & HW-TSC$^\ddag$ & 17.89 & 1.98 & - & - \\
    \multirow{-4}{*}{en-ja} & \textbf{SimulSeamless$^\dag$} & 22.19 & 1.997 (4.018) & 2.137 (4.272) & 0.580 (2.728) \\
    \hline
    & NAIST & 22.11 & 1.471 & 1.907 & 0.668 \\ 
    & CUNI-KIT$^\dag$ & 26.6 & 1.987 (3.508) & - & - \\
    & HW-TSC$^\ddag$ & 27.23 & 1.98 & - & - \\
    & XIAOMI$^\dag$ & 26.59 & 1.966 & - & - \\
    \multirow{-5}{*}{en-zh} & \textbf{SimulSeamless$^\dag$} & 20.56 & 1.942 (3.388) & 2.080 (3.465) & 0.765 (1.933) \\
    \hline
    cs-en & \textbf{SimulSeamless$^\dag$} & 18.03 & 1.988 (3.755) & 2368 (3.999) & 2.778 (3.399) \\
    \hline
  \end{tabular}
  \caption{Results on the MuST-C v2.0 tst-COMMON (for en-\{de, ja, zh\}) and IWSLT 2024 dev (for cs-en) considering BLEU and all the latency metrics (in seconds) reported for the task. Results in brackets are computationally aware but computed with different environments between systems. $^\dag$ indicates systems trained offline and tested in simultaneous. $^\ddag$ indicates cascade systems.}
  \label{tab:submission}
\end{table*}

\subsection{Comparison with Last Year's Participants}
In Table \ref{tab:submission}, we report the scores for the final submission for each language pair, including LAAL and ATD latency metrics and their corresponding computationally aware scores. SimulSeamless is compared with all the participants of last year: CMU \citep{yan-etal-2023-cmus}, CUNI-KIT \citep{polak-etal-2023-towards}, FBK \citep{papi-etal-2023-direct}, HW-TSC \citep{guo-etal-2023-hw}, NAIST \citep{fukuda-etal-2023-naist}, and XIAOMI \citep{huang-etal-2023-xiaomi}. Comparisons are not reported for cs-en since it is a new language direction for the task.

First, it can be noticed that SimulSeamless achieves the best translation quality and, in general, the best quality-latency trade-off for en-ja. Conversely, it struggles to achieve very competitive results in en-de and, especially, in en-zh. However, it is important to notice that SimulSeamless is the only model that has not been fine-tuned on the IWSLT-allowed data for the task, which include the MuST-C v2.0 training set. Therefore, it
is a more generic and multilingual system covering more than 143 source languages and 200 target languages.\footnote{We are not able to exclude that MuST-C has been used for training the "off-the-shelf" SeamlessM4T but no ad-hoc fine-tuning on the data and/or language pairs has been performed for our participation.} 

Furthermore, an overlap has been identified between the MuST-C tst-COMMON and the ST-TED dataset \citep{zhang-ao-2022-yitrans}, which was allowed for last year's task. Some participants, unaware of this issue, employed the ST-TED dataset (e.g., CUNI-KIT and XIAOMI). Therefore, the results achieved by last year's submissions on the MuST-C tst-COMMON may not be entirely reliable.
In addition, it has been recently found another possible overlap with TED2020, which may invalidate other scores.\footnote{Unaware of this overlap, participations from CMU and HW-TSC used this dataset.} 


In conclusion, SimulSeamless allows for acceptable or even better results compared to last year's participants in the SimulST Evaluation Campaign while being generic and potentially applicable to all translation directions supported by the underlying SeamlessM4T model without any retraining or adaptation.



\section{Conclusions}

We introduced FBK's system designed for participation in the IWSLT 2024 Evaluation Campaigns in Simultaneous Translation and, specifically, the speech-to-text sub-track (SimulST). 
Our submission is characterized by the "off-the-self" use of the SeamlessM4T model for direct speech translation, repurposed for the simultaneous scenario by 
means of AlignAtt.
AlignAtt is a SimulST policy that leverages cross-attention scores to guide simultaneous inference without any further modification or adaptation of the underlying model. 
The combination of SeamlessM4T and AlignAtt results in SimulSeamless, which supports all translation pairs of the Evaluation Campaign (English to German, Japanese, and Chinese, and Czech to English). SimulSeamless, to be released upon paper acceptance, achieves acceptable or even superior results compared to last year's participants. Moreover, it can be used for any language pairs enabled by the underlying SeamlessM4T model, potentially covering more than 143 source languages and 200 target languages. 



\section*{Acknowledgments}

The work presented in this paper is funded by the European Union's Horizon research and innovation programme under grant agreement No 101135798, project Meetween (My Personal AI Mediator for Virtual MEETtings BetWEEN People), and the PNRR project FAIR -  Future AI Research (PE00000013), under the NRRP MUR program funded by the NextGenerationEU.

\bibliography{custom}




\end{document}

%% file: de_tst_mustv2.tex
\pgfplotstableread[row sep=\\]{
BLEU	Layer \\
27.29   12 \\
27.37   11 \\ 
27.39   10 \\
27.42   9 \\
27.41   8 \\
27.41   7 \\
27.40   6 \\
27.41   5 \\
27.36   4 \\
27.16   3 \\
27.24   2 \\
27.33   1 \\
}\ende

\pgfplotstableread[row sep=\\]{
BLEU	Layer \\
20.65   12\\
20.63   11\\
20.67   10\\
20.76   9\\
20.68   8\\
20.74   7\\
20.63   6\\
20.54   5\\
20.55   4\\
20.34   3\\
20.45   2\\
20.22   1\\
}\enzh

\pgfplotstableread[row sep=\\]{
BLEU	Layer \\
22.27   12\\
22.48   11\\
22.37   10\\
22.41   9\\
22.43   8\\
22.47   7\\
22.35   6\\
22.36   5\\
22.41   4\\
22.24   3\\
22.10   2\\
22.19   1\\
}\enja

\pgfplotstableread[row sep=\\]{
BLEU	Layer \\
17.87   12\\
17.84   11\\
17.65   10\\
17.93   9\\
17.99   8\\
17.84   7\\
17.81   6\\
18.13   5\\
17.88   4\\
17.89   3\\
17.85   2\\
18.03   1\\
}\csen

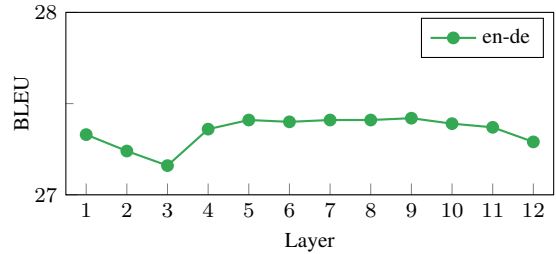
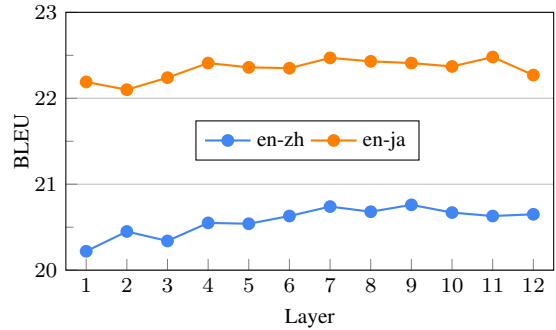
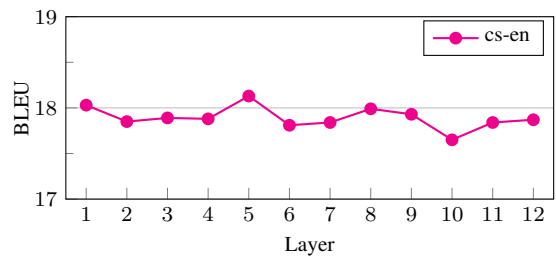
\begin{figure}[!ht]
\centering
\small
\subfigure[English to German]{
\begin{tikzpicture}
    \begin{axis}[
            ymajorgrids=true,
            xtick pos=left,
            ytick pos=left,
            minor y tick num=1,
            minor x tick num=0,
            ytick={27,28},
            ymin=27,
            ymax=28,
            xmin=0.5,
            xmax=12.5,
            ylabel=BLEU, xlabel=Layer,
            ylabel shift={-4pt},
            width=8cm,
            height=4cm,
            xtick=data,
            compat=newest,
            xtick={1,2,3,4,5,6,7,8,9,10,11,12},
            every axis plot/.append style={thick},
            legend style={at={(0.98,0.75)},anchor=south east,legend columns=2},
        ]
        \addplot[color=darkgreen, mark=*] table[x=Layer,y=BLEU]{\ende};
    \legend{en-de}
    \end{axis}
\end{tikzpicture}
}
\subfigure[English to Chinese and Japanese]{
\begin{tikzpicture}
    \begin{axis}[
            ymajorgrids=true,
            xtick pos=left,
            ytick pos=left,
            minor y tick num=1,
            minor x tick num=0,
            ytick={20,21,22,23},
            ymin=20,
            ymax=23,
            xmin=0.5,
            xmax=12.5,
            ylabel=BLEU, xlabel=Layer,
            ylabel shift={-4pt},
            width=8cm,
            height=5cm,
            xtick=data,
            compat=newest,
            xtick={1,2,3,4,5,6,7,8,9,10,11,12},
            every axis plot/.append style={thick},
            legend style={at={(0.725,0.425)},anchor=south east,legend columns=2},
        ]
        \addplot[color=lightblue, mark=*] table[x=Layer,y=BLEU]{\enzh};
        \addplot[color=orange, mark=*] table[x=Layer,y=BLEU]{\enja};
    \legend{en-zh,en-ja}
    \end{axis}
\end{tikzpicture}
}
\subfigure[Czech to English]{
\begin{tikzpicture}
    \begin{axis}[
            ymajorgrids=true,
            xtick pos=left,
            ytick pos=left,
            minor y tick num=1,
            minor x tick num=0,
            ytick={17,18,19},
            ymin=17,
            ymax=19,
            xmin=0.5,
            xmax=12.5,
            ylabel=BLEU, xlabel=Layer,
            ylabel shift={-4pt},
            width=8cm,
            height=4cm,
            xtick=data,
            compat=newest,
            xtick={1,2,3,4,5,6,7,8,9,10,11,12},
            every axis plot/.append style={thick},
        ]
        \addplot[color=magenta, mark=*] table[x=Layer,y=BLEU]{\csen};
    \legend{cs-en}
    \end{axis}
\end{tikzpicture}
}
\caption{Translation quality (BLEU$\uparrow$) scores of SimulSeamless on MuST-C v2.0 tst-COMMON for English (en) to German (de), Japanese (ja), and Chinese (zh), and on the IWSLT 2024 dev set for Czech (cs) to English by varying the decoder layer from which cross-attention scores are extracted from.}
\label{fig:bleu_res}
\end{figure}

\pgfplotstableread[row sep=\\]{
AL	Layer \\
2.088   12\\
2.058   11\\
2.061   10\\
2.039   9\\
2.025   8\\
2.013   7\\
2.044   6\\
1.911   5\\
1.815   4\\
1.821   3\\
1.804   2\\
1.924   1\\
}\ende

\pgfplotstableread[row sep=\\]{
AL	Layer \\
2.045   12\\
2.017   11\\
2.023   10\\
2.028   9\\
2.014   8\\
2.016   7\\
2.017   6\\
1.959   5\\
1.942   4\\
1.922   3\\
1.934   2\\
1.927   1\\
}\enzh

\pgfplotstableread[row sep=\\]{
AL	Layer \\
2.213   12\\
2.189   11\\
2.244   10\\
2.195   9\\
2.207   8\\
2.200   7\\
2.176   6\\
2.055   5\\
2.041   4\\
2.015   3\\
1.963   2\\
1.997   1\\
}\enja

\pgfplotstableread[row sep=\\]{
AL	Layer \\
2.258    12\\
2.308    11\\
2.339    10\\
2.284    9\\
2.269    8\\
2.200    7\\
2.270    6\\
2.111    5\\
1.962    4\\
1.970    3\\
1.923    2\\
1.988    1\\
}\csen

\pgfplotstableread[row sep=\\]{
AL	Layer \\
2	0 \\
2   13 \\
}\thr

\begin{figure}[!ht]
\centering
\small
\begin{tikzpicture}
    \begin{axis}[
            ymajorgrids=true,
            xtick pos=left,
            ytick pos=left,
            minor y tick num=1,
            minor x tick num=0,
            ytick={1.5,2,2.5},
            extra y ticks={1.7,1.8,1.9,2.1,2.2,2.3,2.4},
            ymin=1.75,
            ymax=2.35,
            xmin=0.5,
            xmax=12.5,
            ylabel=AL (s), xlabel=Layer,
            ylabel shift={-4pt},
            width=8cm,
            height=6cm,
            xtick=data,
            compat=newest,
            xtick={1,2,3,4,5,6,7,8,9,10,11,12},
            every axis plot/.append style={thick},
            legend style={at={(0.96,0.06)},anchor=south east,legend columns=1},
        ]
        \addplot[color=darkgreen, mark=*] table[x=Layer,y=AL]{\ende};
        \addplot[color=lightblue, mark=*] table[x=Layer,y=AL]{\enzh};
        \addplot[color=orange, mark=*] table[x=Layer,y=AL]{\enja};
        \addplot[color=magenta, mark=*] table[x=Layer,y=AL]{\csen};
        \addplot[dashed, color=red, mark=] table[x=Layer,y=AL]{\thr};
        \legend{en-de,en-zh,en-ja,cs-en}
    \end{axis}
\end{tikzpicture}
\caption{Latency (AL$\downarrow$) scores of SimulSeamless on MuST-C v2.0 tst-COMMON for English (en) to German (de), Japanese (ja), and Chinese (zh), and on the IWSLT 2024 dev set for Czech (cs) to English by varying the decoder layer from which cross-attention scores are extracted from.}
\label{fig:al_res}
\end{figure}
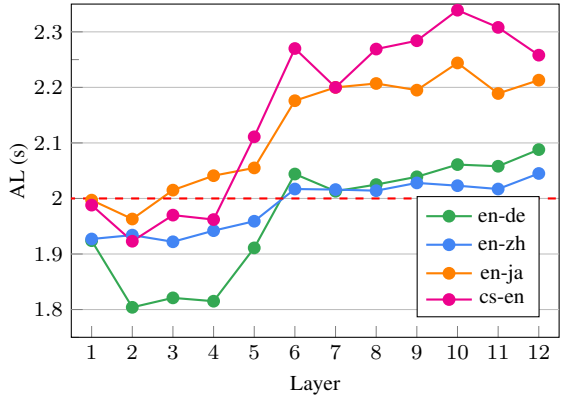